\documentclass[10pt,twocolumn,letterpaper]{article}

\usepackage{wacv}
\usepackage{times}
\usepackage{epsfig}
\usepackage{graphicx}
\usepackage{amsmath}
\usepackage{amssymb}
\usepackage{booktabs}

\def\wacvPaperID{981} 

\wacvapplicationstrack 

\wacvfinalcopy 

\usepackage{graphics} 
\usepackage{epsfig} 
\usepackage{mathptmx} 
\usepackage{times} 
\usepackage{amsmath} 
\usepackage{amssymb}  
\usepackage{multirow}
\usepackage{booktabs}
\usepackage{float}
\usepackage{threeparttable}
\usepackage{bm}

\usepackage{cite}

\usepackage{color, soul}
\sethlcolor{blue}

\newcommand{\hlb}[1]{{\color{blue}{#1}}}

\ifwacvfinal
\usepackage[breaklinks=true,bookmarks=false]{hyperref}
\else
\usepackage[pagebackref=true,breaklinks=true,colorlinks,bookmarks=false]{hyperref}
\fi

\begin{document}

\title{BEVSegFormer: Bird's Eye View Semantic Segmentation From Arbitrary Camera Rigs}

\author{Lang Peng\textsuperscript{\rm 1}, 
        Zhirong Chen\textsuperscript{\rm 1}, 
        Zhangjie Fu\textsuperscript{\rm 1}, 
        Pengpeng Liang\textsuperscript{\rm 2}, 
        Erkang Cheng\thanks{Corresponding author.}~~\textsuperscript{\rm 1}
 \\
 \textsuperscript{\rm 1}Nullmax~~~~~~\textsuperscript{\rm 2}Zhengzhou University\\
 {\tt\small \{penglang, chenzhirong, fuzhangjie,chengerkang\}@nullmax.ai,~liangpcs@gmail.com}
}

\maketitle
\thispagestyle{empty}

\begin{abstract}
Semantic segmentation in bird's eye view (BEV) is an important task for autonomous driving. Though this task has attracted a large amount of research efforts, it is still challenging to flexibly cope with arbitrary (single or multiple) camera sensors equipped on the autonomous vehicle. In this paper, we present BEVSegFormer, an effective transformer-based method for BEV semantic segmentation from arbitrary camera rigs. Specifically, our method first encodes image features from arbitrary cameras with a shared backbone. These image features are then enhanced by a deformable transformer-based encoder. Moreover, we introduce a BEV transformer decoder module to parse BEV semantic segmentation results. An efficient multi-camera deformable attention unit is designed to carry out the BEV-to-image view transformation. Finally, the queries are reshaped according to the layout of grids in the BEV, and upsampled to produce the semantic segmentation result in a supervised manner. We evaluate the proposed algorithm on the public nuScenes dataset and a self-collected dataset. Experimental results show that our method achieves promising performance on BEV semantic segmentation from arbitrary camera rigs. We also demonstrate the effectiveness of each component via ablation study.
\end{abstract}

\section{Introduction}

Bird's-eye-view (BEV) representation of perception information is critical in the autonomous driving or robot navigation system as it is convenient for planning and control tasks. For example, in a mapless navigation solution, building a local BEV map provides an alternative to High-definition map (HD map) and is important for down-streaming tasks of the perception system including behavior prediction of agents and motion planning. BEV semantic segmentation from cameras is usually treated as the first step to build the local BEV map.

\begin{figure}[t]
  \centering{\includegraphics[width=0.9\linewidth]{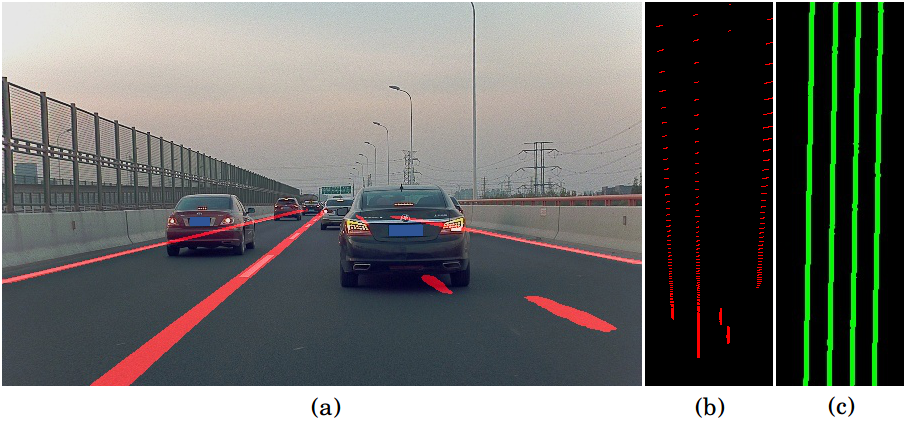}}\\
  \caption{Comparison of lane segmentation results on image space and BEV space. (a) lane segmentation on image space, (b) BEV segmentation by IPM view transformation of (a), (c) our BEV lane segmentation.} 
  \label{figure:uv_vs_bev}
\end{figure}

To obtain BEV semantic segmentation from cameras, traditional methods usually generate segmentation results in the image space and then convert it  to the BEV space by inverse perspective mapping (IPM) function. Although IPM is a straightforward and simple way to bridge the image space and the BEV space, it requires accurate intrinsic and extrinsic parameters of the camera or real-time camera pose estimation. Therefore, it is likely to produce inferior view transformation. 
Taking lane segmentation as an example,  as shown in Fig.~\ref{figure:uv_vs_bev}, traditional methods with IPM provide inaccurate results in challenging scenarios where  occlusions are present or at the distant areas.

Recently, deep learning approaches have been studied for BEV semantic segmentation~\cite{reiher2020sim2real,philion2020lift,pan2020cross,roddick2020predicting,hu2021fiery, ng2020bev,schulter2018learning}.  Lift-Splat-Shoot~\cite{philion2020lift} completes the view transformation from image view to BEV with pixel-wise depth estimation results. Using depth estimation increases the complexity of the view transformation process.
Some approaches apply MLP~\cite{pan2020cross} or FC~\cite{roddick2020predicting} operators to perform the view transformation. 
These fixed view transformation methods learn a fixed mapping between the image space and the BEV space, and therefore are not dependent on the input data.

Transformer-based approaches are another line of research for perception in BEV space. In object detection task, DETR3D~\cite{wang2021detr3d} introduces a 3D bounding boxes detection method that directly generates predictions in 3D space from 2D features of multiple camera images. The view transformation between 3D space and 2D image space is achieved by 3D-to-2D sparse queries of a cross-attention module.

\begin{figure}[t]
  \centering{\includegraphics[width=0.9\linewidth]{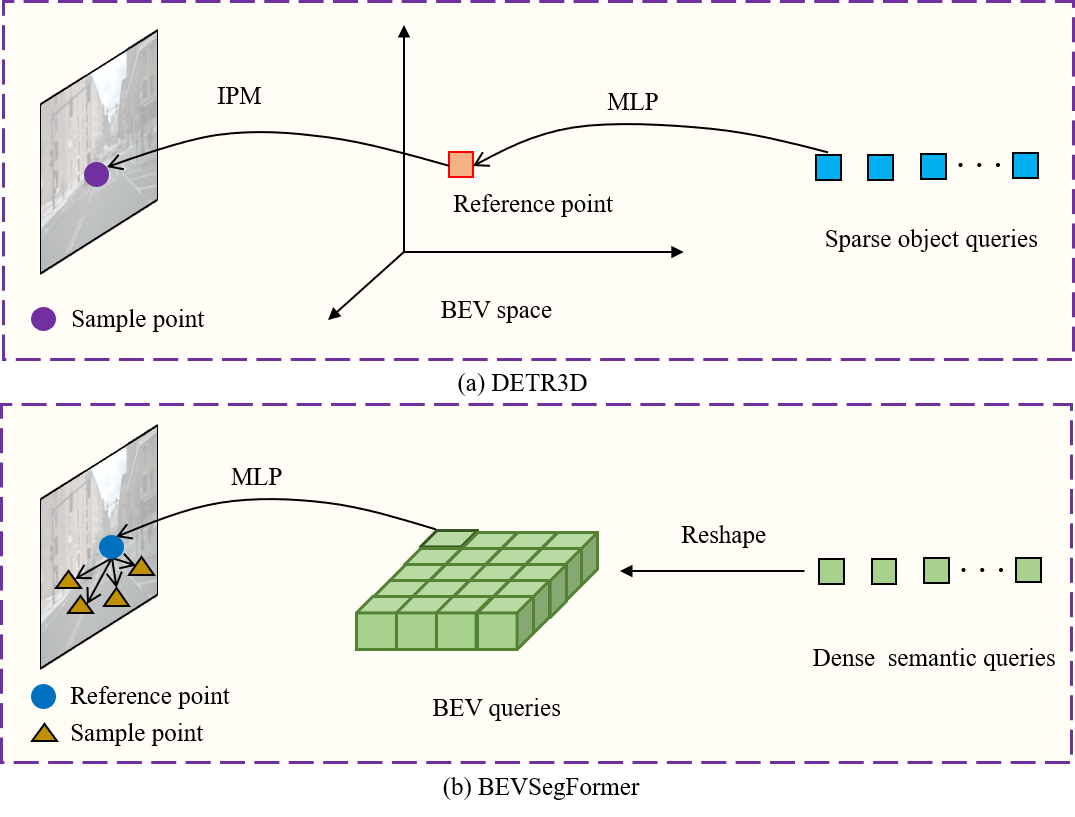}}\\
  \caption{Comparison of view transformation between our BEV-to-image scheme and 3D-to-2D method in DETR3D~\cite{wang2021detr3d}.} 
  \label{figure:view_trans}
\end{figure}

Inspired by DETR3D~\cite{wang2021detr3d}, we propose a method to compute the view transformation by BEV-to-image queries using the cross-attention mechanism in the transformer. With the novel view transformation approach, we build a BEV semantic segmentation method, BEVSegFormer, to perform BEV semantic segmentation from arbitrary camera configurations. As shown in Fig.~\ref{figure:view_trans}, our approach differs from DETR3D~\cite{wang2021detr3d} in several ways. First, DETR3D constructs a set of sparse object queries, while our BEVSegFormer builds dense BEV queries for the semantic segmentation task. Second, in DETR3D, a query applies MLP to predict a reference point in 3D space and then projects it back to image feature space by IPM which requires camera extrinsic parameters. In contrast, a query in our method directly predicts a reference point on image feature space through MLP operator. In this way, the view transformation of our method does not rely on camera extrinsic parameters. In addition, in order to encode more image context features for a query, our method also uses deformable attention to regress sampled points around the reference point.

\begin{figure*}[t]
  \centering{\includegraphics[width=0.95\linewidth]{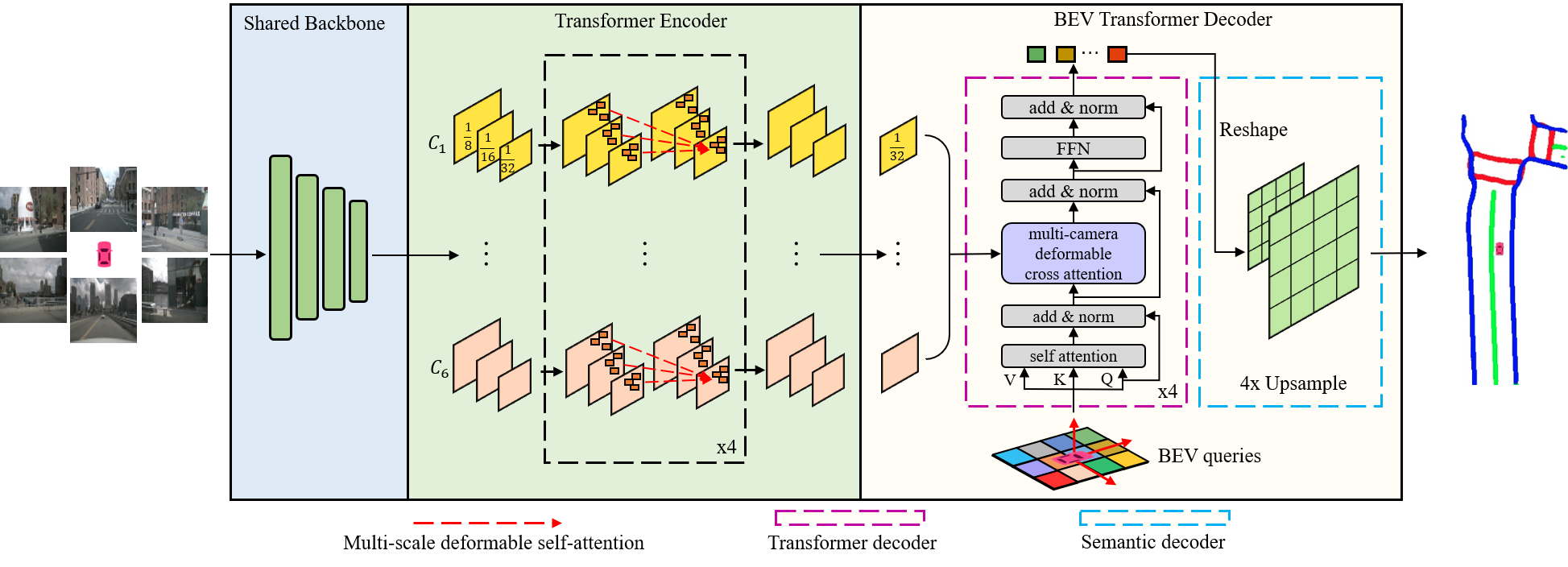}}\\
  \caption{Overview of the proposed network. Our method consists of a Shared Backbone, a Transformer Encoder and a BEV Transformer Decoder module. }
  \label{figure:overview}
\end{figure*}

Our BEVSegFormer consists of three major components: (1) a shared backbone to extract feature maps of arbitrary cameras; (2) a transformer encoder to embed feature maps by a self-attention module; and (3) a BEV transformer decoder to process BEV queries by cross-attention mechanism and output the final BEV semantic segmentation result. In the BEV transformer decoder, we introduce a multi-camera deformable cross-attention module to link feature maps from multiple cameras to BEV queries, without the need of camera intrinsic and extrinsic parameters. Specifically, a BEV query pays more attention to the learned reference points and the corresponding sampled points in multi-camera features. In this way, the multi-camera deformable cross-attention module provides an effective method to complete the BEV-to-image view transformation.

In the experiments, we evaluate our BEVSegFormer on the public nuScenes dataset and a self-collected dataset. Our proposed BEVSegFormer sets a new state-of-the-art for BEV segmentation on the nuScenes validation set without using temporal information. The effectiveness of each proposed component is validated as well.

\section{RELATED WORK}

\textbf{Semantic Segmentation.} Semantic segmentation plays a vital role in high-level scene understanding and is a fundamental problem in computer vision. With the rapid development of CNNs, recent methods like FCN~\cite{shelhamer2016fully} and U-Net~\cite{ronneberger2015u} apply the encoder-decoder architecture to learn the dense prediction with feature extraction backbones (e.g., VGGNet~\cite{simonyan2014very} and ResNet~\cite{he2016deep}). Existing methods also make use of strategies to further improve the segmentation performance, including atrous convolution~\cite{chen2017deeplab}, pyramid pooling module~\cite{zhao2017pyramid}, chained residual pooling mechanism~\cite{lin2017refinenet} and so on. In this paper, we apply ResNet as part of the encoder to extract semantic features.

\textbf{BEV Semantic Segmentation.} In autonomous driving and robot navigation, BEV semantic segmentation is an important perception task for downstream functions, such as behavior prediction and planning. 
Cam2BEV~\cite{reiher2020sim2real} performs a spatial transformer module to transform perspective features to BEV space from surrounding inputs by IPM, which is a straightforward way to link image space to BEV under flat ground assumption. SBEVNet~\cite{gupta2022sbevnet} first uses stereo image features to generate disparity feature volume to enhance the BEV representation, and then estimates the BEV semantic layout on the BEV representation via a U-Net~\cite{ronneberger2015u} model. Methods in~\cite{schulter2018learning, philion2020lift, hu2021fiery, ng2020bev} utilize depth information to perform the view transformation. For example, Lift-Splat-Shoot~\cite{philion2020lift} first estimates implicit pixel-wise depth information and then uses camera geometry to build the connection between BEV segmentation and feature maps. FIERY~\cite{hu2021fiery} takes a similar approach for video based input. BEV-Seg~\cite{ng2020bev} uses an additional parse network to refine the projected and incomplete BEV segmentation by depth result. VPN~\cite{pan2020cross} uses MLP to directly generate BEV segmentation from cross-view images. FishingNet~\cite{hendy2020fishing} extends it to support multiple sensors. PyrOccNet~\cite{roddick2020predicting} applies FC operators to complete the transformation at different distance ranges. Similarly, HDMapNet~\cite{li2021hdmapnet} also applies FC operators to build the local BEV map from multiple cameras. Different from previous works, Monolayout~\cite{mani2020monolayout} performs view transformation via a standard encoder-decoder structure.

\textbf{Transformer-based Semantic Segmentation.} Transformer, first applied in the field of natural language processing, has been widely used in many other computer vision tasks. For example, ViT~\cite{dosovitskiy2020image} introduces transformer encoder for image classification task. 
DETR~\cite{carion2020end} and its variant~\cite{zhu2020deformable} are proposed for object detection by a transformer encoder-decoder architecture. 
SETR~\cite{zheng2021rethinking} is the first to expand transformer encoder for segmentation task.
SegFormer~\cite{xie2021segformer} proposes an efficient semantic segmentation framework, which combines a hierarchical  transformer encoder with a lightweight MLP decoder. 
FTN~\cite{wu2021fully} introduces a full transformer encoder and decoder network for image segmentation.
Adapted from DETR~\cite{carion2020end}, MaskFormer~\cite{cheng2021per} employs a transformer decoder to compute a set of pairs, each consisting of a class prediction and a mask embedding vector which is then combined with pixel-wise embedding from a FCN. It offers both semantic and instance segmentation in a unified manner.

Recently, transformer has been used to perform view transformation. 
For example, NEAT~\cite{chitta2021neat} uses transformer encoder and converts image features to BEV space with a MLP-based attention by traversing through all the grid in the BEV space. 
PYVA~\cite{yang2021projecting} generates BEV features from a CNN encoder-decoder structure by MLP. The feature is then enhanced by a transformer cross-attention module. 
Method in~\cite{saha2021translating} introduces an encoder-decoder transformer block to column-wisely translate spatial features from the image to BEV. BEVFormer~\cite{li2022bevformer} transfers a set of 3D points back to image space and uses deformable attention to build a dense BEV  map from image features. The projection is computed by camera extrinsic parameters. BEVerse~\cite{zhang2022beverse} takes method in Lift-Splat-Shoot~\cite{philion2020lift} which performs view transformation by depth estimation with camera geometry. PETR~\cite{liu2022petr, liu2022petrv2} brings a new perspective of view transformation which encodes 3D coordinate information into 2D image features. The transformer is then applied to connect the BEV space and image space by the 3D information boosted image features. Temporal information and multiple downstream tasks are also investigated in ~\cite{li2022bevformer, zhang2022beverse, liu2022petrv2}. In this paper, we mainly focus on spatial segmentation task in BEV space. Our method has the potential to extend to associate temporal information and apply to other downstream perception tasks.

Different with these works, we introduce a BEV transformer decoder module to parse BEV semantic segmentation from the image features. 
The paradigm of view transformation between BEV space and image space does not require camera extrinsic parameters and it naturally supports arbitrary camera settings.
Specifically, an efficient multi-camera deformable cross attention unit is designed to carry out the BEV-to-image view transformation. The queries are reshaped according to the layout of grids in the BEV, and upsampled to produce the semantic segmentation result in a supervised manner.

\section{Method}
Fig.~\ref{figure:overview} shows the overview of our BEVSegFormer method. It consists of three parts: (1) a Shared Backbone to process arbitrary cameras and output feature maps; (2) a Transformer Encoder to enhance the feature representation and (3) a BEV Transformer Decoder to process BEV queries by cross-attention mechanism and then parses the output queries to BEV semantic segmentation.

\subsection{Shared Backbone}

For a single input image, the backbone takes the input and outputs multi-scale feature maps. For multiple camera configurations, these multiple images share the same backbone and output corresponding feature maps. We take ResNet as the backbone in the experiments.

\subsection{Transformer Encoder}

In the transformer encoder, we first apply $1\times1$ conv operators on $c3$, $c4$, $c5$ stage features from the shared backbone to get multi-scale features $ \left \{ \bm{x}^{l} \right \}_{l=1}^{L}$, where ${\bm{x}}^{l} \in \mathbb{R}^{C_l\times H_l\times W_l}$, $C_l$ is channel number of feature map, $H_l$ and $W_l$ denotes the height and width of the feature map at $l$-th scale. Similar to~\cite{zhu2020deformable}, we apply multi-scale deformable self attention module separately on feature maps generated by each scale, and create additional learnable scale level position embedding for each scale feature map. Multi-scale deformable self attention module does not require to compute the dense attention map and only focuses on a set of sampling points near a reference point. The transformer encoder outputs an enhanced multi-scale features for each camera.

\subsection{BEV Transformer Decoder}

\begin{figure}[t]
  \centering
  \includegraphics[width=0.95\linewidth]{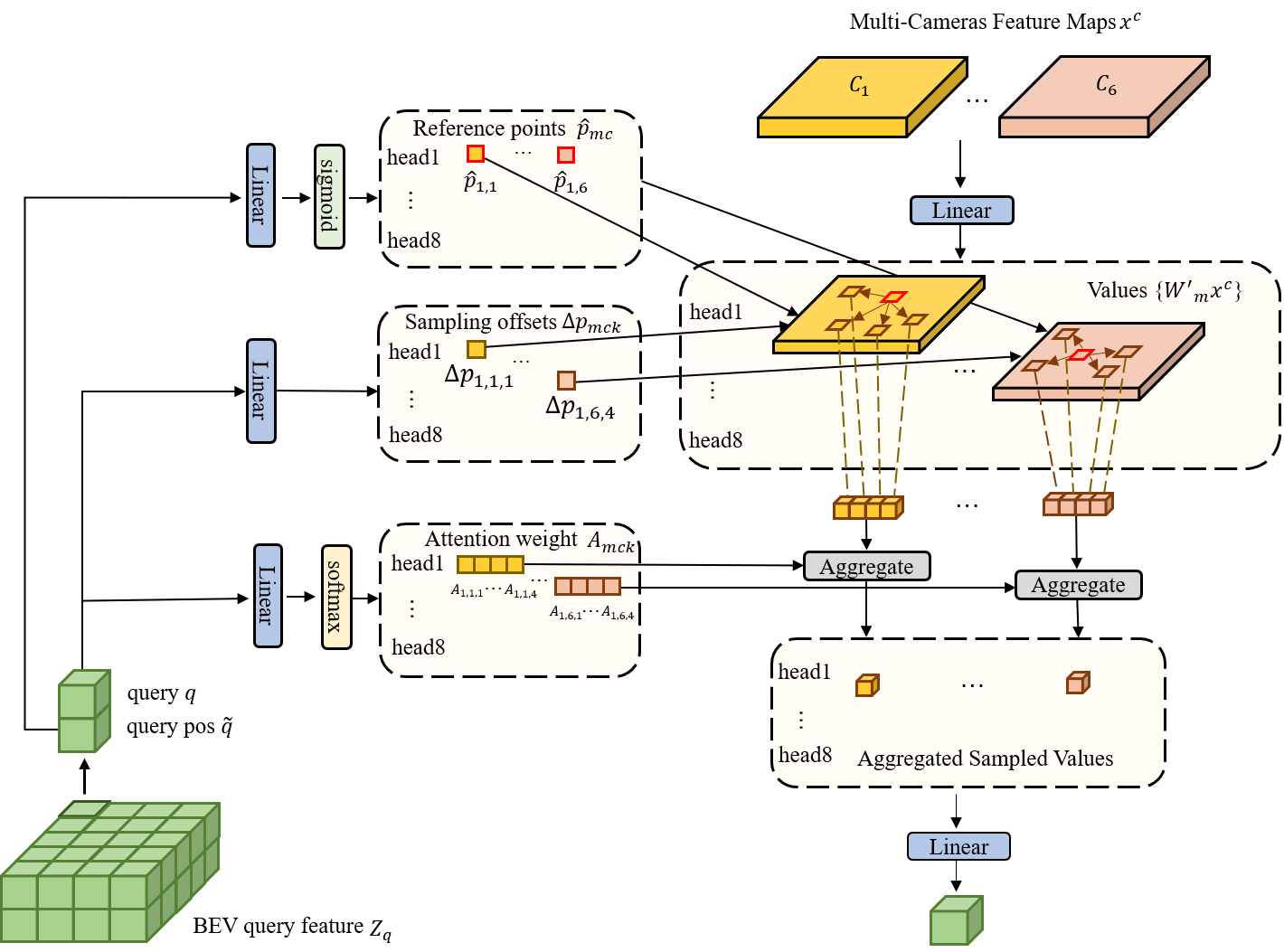}\\
  \caption{Illustration of the Multi-Camera Deformable Cross-Attention module in BEV Transformer Decoder.} 
  \label{figure:cross_attention}
\end{figure}

Our BEV transformer decoder includes a transformer decoder to compute cross-attention between BEV queries and multi-camera feature maps, and a semantic decoder to parse queries to BEV segmentation result.

In transformer decoder, we construct $H_q \times W_q$ queries on 2D BEV space and these BEV queries are then treated as $N_q = H_q \times W_q$ regular queries in the cross-attention module. The dense BEV query embeddings are denoted as $\bm{z}_q \in \mathbb{R}^{C \times N_q  }$. We only use the smallest resolution (1/32 of original input resolution) of the multi-scale feature maps as input of the transformer decoder. 

We adapt deformable attention module in deformable DETR~\cite{zhu2020deformable} to a multi-camera deformable cross-attention module which is able to transform feature maps from multiple cameras to BEV queries, without the need of camera intrinsic and extrinsic parameters. Mathematically, let $q$ be a query element in $\bm{z}_q$ and its reference point $\hat{\bm{p}}_q$, the multi-camera deformable attention module is written as:

\begin{equation}
\begin{aligned}
 &\text{MultiCameraDeformAttn}(\bm{z}_q,  \hat{p}_q ,\left \{ x^{c} \right \}_{c=1}^{N_c})  =  \\
&\sum_{m=1}^{M} \bm{W}_m [ \sum_{c=1}^{N_c} \sum_{k=1}^{K} A_{mcqk} \cdot \bm{W}_{m}^{'} \bm{x^{c}}(\phi_c( \bm{\hat{p}}_q )  + \Delta \bm{P}_{mcqk})] ,
\end{aligned}
\label{mcda}
\end{equation}
where $(m,c,k)$ indicates the index of the attention head, camera, and sampling point, respectively. $\bm{W}_m \in \mathbb{R}^{C_v \times C  }$and $\bm{W}_{m}^{'} \in \mathbb{R}^{C\times C_v}$ are the learnable parameter matrices of the linear projection layer, $C=C_v \times M $ by default. $\Delta \bm{P}_{mcqk}$ and $A_{mcqk}$ are the sampling offset and attention weight of the $k$-th sampling point in the $c$-th camera  and $m$-th attention head. The scalar attention weight $A_{mcqk}$ is normalized to sum as 1. $\phi_c( \bm{\hat{\bm{p}}}_q )$ re-scales the normalized coordinates $\bm{\hat{p}}_q$ to the input feature map.

Fig.~\ref{figure:cross_attention} shows the overall structure of the multi-camera deformbale attention. For each BEV query $q$, we apply a learnable linear projection layer on its position embedding to obtain the 2D coordinate of reference points $\bm{\hat{p}}_q  \in \mathbb{R}^{M\times N_c  \times 2}$, and then use the sigmoid function to normalize these coordinates. Two learnable linear projection layers are used to predict the offset of the sampling points relative to the reference point, and the attention weight of these sampling points. Finally, the camera features at the sampled positions are aggregated by the attention weights to generate a new query. Different from shared reference points in multi-scale features in Deformable DETR~\cite{zhu2020deformable}, we learn independent reference points on multi-camera feature maps, so that the network can automatically select different locations of reference points on multi-camera features.

\subsection{BEV Semantic Decoder}

\begin{figure}[t]
  \centering
  \includegraphics[width=0.9\linewidth]{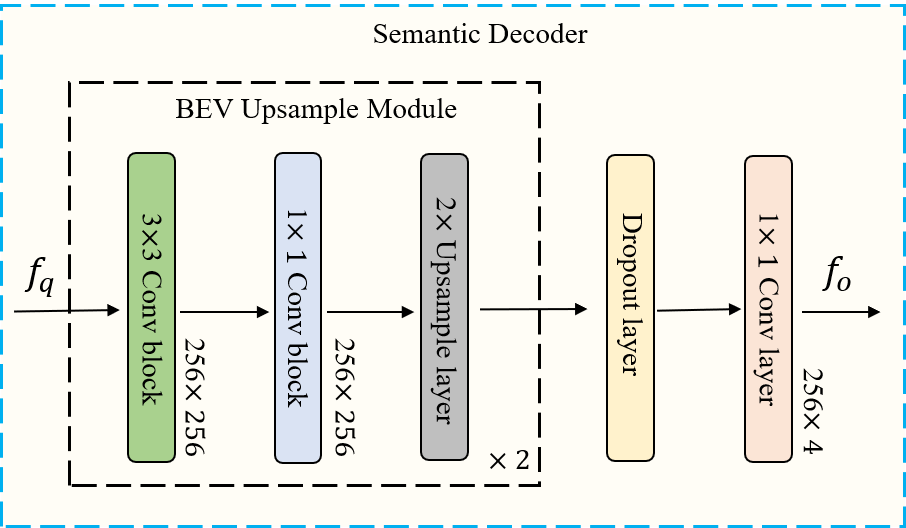}\\
  \caption{Illustration of Semantic Decoder.} 
  \label{figure:semantic_decoder}
  \vspace{-5mm}
\end{figure}

Fig.~\ref{figure:semantic_decoder} illustrates the BEV semantic decoder architecture. In semantic decoder, we reshape the BEV query features $\bm{z}_q \in \mathbb{R}^{C \times N_q }$ from the transformer decoder into a 2D spatial feature $\bm{f_q} \in \mathbb{R}^{C \times H_q  \times W_q }$. The 2D spatial feature $\bm{f_q}$ is then fed to a two-stage BEV Upsample Module, a dropout layer and a $1 \times 1$ convolution layer to compute the final output $\bm{f_o} \in \mathbb{R}^{C_{seg} \times H_q  \times W_q }$. Each stage of the BEV Upsample Module consists of a $3\times3$ convolution block, a $1\times1$ convolution block and a $2\times$ bilinear interpolation operation. A convolutional block includes a convolution layer, a BN layer and a ReLU activation function. The dropout layer with a dropout rate of 0.1 is applied to the feature map after restoring the resolution.

\section{Experiment}

\begin{table*}
\caption{Comparison to state of the art methods on nuScenes dataset without using temporal information. ‘\dag’ are the results reported in HDMapNet~\cite{li2021hdmapnet}. }
 \begin{center}
 \begin{small}
\begin{tabular}{c|cccc}
\hline
  \multicolumn{1}{c|}{\multirow{2}{*}{Method}} 
& \multicolumn{4}{c}{\multirow{1}{*}{IoU}} \\
  \multicolumn{1}{c|}{}  
& \multicolumn{1}{c}{Divider}  
& \multicolumn{1}{c}{Ped Crossing} 
& \multicolumn{1}{c}{Boundary} 
& \multicolumn{1}{c}{All}          \\

\hline
IPM (B)\dag                                 & 25.5           & 12.1           & 27.1           & 21.6           \\
IPM (BC)\dag                                & 38.6           & 19.3           & 39.3           & 32.4           \\
Lift-Splat-Shoot\dag~\cite{philion2020lift}                       & 38.3           & 14.9           & 39.3           & 30.8           \\
VPN\dag~\cite{pan2020cross}                                    & 36.5           & 15.8           & 35.6           & 29.3           \\
HDMapNet (Surr)\dag~\cite{li2021hdmapnet}   & 40.6           & 18.7           & 39.5           & 32.9           \\
\hline
\hline
BEVSegFormer                           & \textbf{51.08} & \textbf{32.59} & \textbf{49.97} & \textbf{44.55} \\

\hline
  \end{tabular}
  \end{small}
 \end{center}
\label{tab:sota_nuscene}
\end{table*}

\begin{table}[!t]
\caption{Result of BEV segmentation of front camera on nuScenes dataset.}
\begin{center}
  \begin{small}
\begin{tabular}{c|cccc}
\hline
  \multicolumn{1}{c}{\multirow{2}{*}{\#Queries}} 
& \multicolumn{4}{|c}{\multirow{1}{*}{IoU}} \\
& \multicolumn{1}{c}{Divider}  
& \multicolumn{1}{c}{Ped Crossing} 
& \multicolumn{1}{c}{Boundary} 
& \multicolumn{1}{c}{All}   \\
\hline
1250    & 34.56 &  15.41 & 30.07  & 26.68           \\
5000    & 37.54 &  17.82 & 34.25  & 29.87           \\
\hline
\end{tabular}
  \end{small}
 \end{center}
 \label{tab:fc_nuscene} 
\end{table}

\begin{table}[!t]
\caption{Result of BEV segmentation on Nullmax Front Camera dataset.}
\begin{center}
  \begin{small}
\begin{tabular}{c|ccc}
\hline
\multicolumn{1}{c}{\multirow{2}{*}{ Method}} 
& \multicolumn{3}{|c}{\multirow{1}{*}{IoU}} \\
& \multicolumn{1}{c}{Background} 
& \multicolumn{1}{c}{Lane} 
& \multicolumn{1}{c}{All}   \\

\hline
M1  &  92.88 & 67.04 & 79.96           \\
\hline
\end{tabular}
  \end{small}
 \end{center}
 \label{tab:fc_nullmax} 
\end{table}

\begin{table*}[!t]
\caption{Results of ablation study for the components of BEVSegFormer on the nuScenes val set. ‘Enc’ denotes encoder of model, ‘Dec’ denotes decoder of model, ‘S’ denotes standard multi-head self attention module and standard multi-head cross attention module, ‘D’ denotes multi-scale deformable self attention module and multi-camera deformable cross attention module. ‘\#Enc’ and ‘\#Dec’denotes number of encoder block and decoder block. ‘CE’ denotes camera position embedding. }
\begin{center}
  \begin{small}
\begin{tabular}{ccccccc|cccc}

\hline
  \multicolumn{1}{c}{\multirow{2}{*}{Method}} 
&\multicolumn{1}{c}{\multirow{2}{*}{Backbone}} 
& \multicolumn{1}{c}{\multirow{2}{*}{Enc}} 
& \multicolumn{1}{c}{\multirow{2}{*}{Dec}} 
& \multicolumn{1}{c}{\multirow{2}{*}{\#Enc}} 
& \multicolumn{1}{c}{\multirow{2}{*}{\#Dec}}
& \multicolumn{1}{c}{\multirow{2}{*}{CE}} 
& \multicolumn{4}{|c}{\multirow{1}{*}{IoU}}   \\
  &  &  &  &  &  & 
& \multicolumn{1}{c}{Divider}
& \multicolumn{1}{c}{Ped Crossing} 
& \multicolumn{1}{c}{Boundary} 
& \multicolumn{1}{c}{All} \\

\hline
M1 & ResNet-34 & S & S & 2 & 2 &            & 44.23 (+0.00)        & 24.32 (+0.00)        & 42.25 (+0.00)         & 36.93 (+0.00)          \\
M2 & ResNet-34 & S & S & 2 & 2 & \checkmark & 45.43 (\hlb{+1.20})  & 26.57 (\hlb{+2.25})  & 45.30 (\hlb{+3.05})   & 39.10 (\hlb{+2.17})    \\
\hline
\hline
M3 & ResNet-34 & S & D & 2 & 2 &            & 47.84 (\hlb{+3.61})  & 28.88 (\hlb{+4.56})  & 46.74 (\hlb{+4.49})   & 41.15 (\hlb{+4.22})    \\
\hline
\hline
M4 & ResNet-34 & D & D & 2 & 2 &            & 49.03 (\hlb{+4.80})  & 30.55 (\hlb{+6.23})  & 48.05 (\hlb{+5.80})   & 42.54 (\hlb{+5.61})    \\
M5 & ResNet-34 & D & D & 4 & 4 &            & 50.79 (\hlb{+6.56})  & 32.39 (\hlb{+8.07})  & 49.84 (\hlb{+7.59})   & 44.34 (\hlb{+7.41})     \\

M6 & ResNet-101 & D & D & 4 & 4 &           & \textbf{51.08} (\hlb{+6.85})  & \textbf{32.59} (\hlb{+8.27})  &         49.97 (\hlb{+7.72})   & \textbf{44.55} (\hlb{+7.62})     \\
M7 & ResNet-101 & D & D & 4 & 4 &\checkmark & 50.29 (\hlb{+6.06})  & 31.82 (\hlb{+7.50})  & \textbf{50.18} (\hlb{+7.93})   & 44.10 (\hlb{+7.17})     \\   
\hline
  \end{tabular}
  \end{small}
 \end{center}
\label{tab:ablation}
\vspace{-3mm}
\end{table*}

\begin{figure}[t]
  \centering{\includegraphics[width=0.95\linewidth]{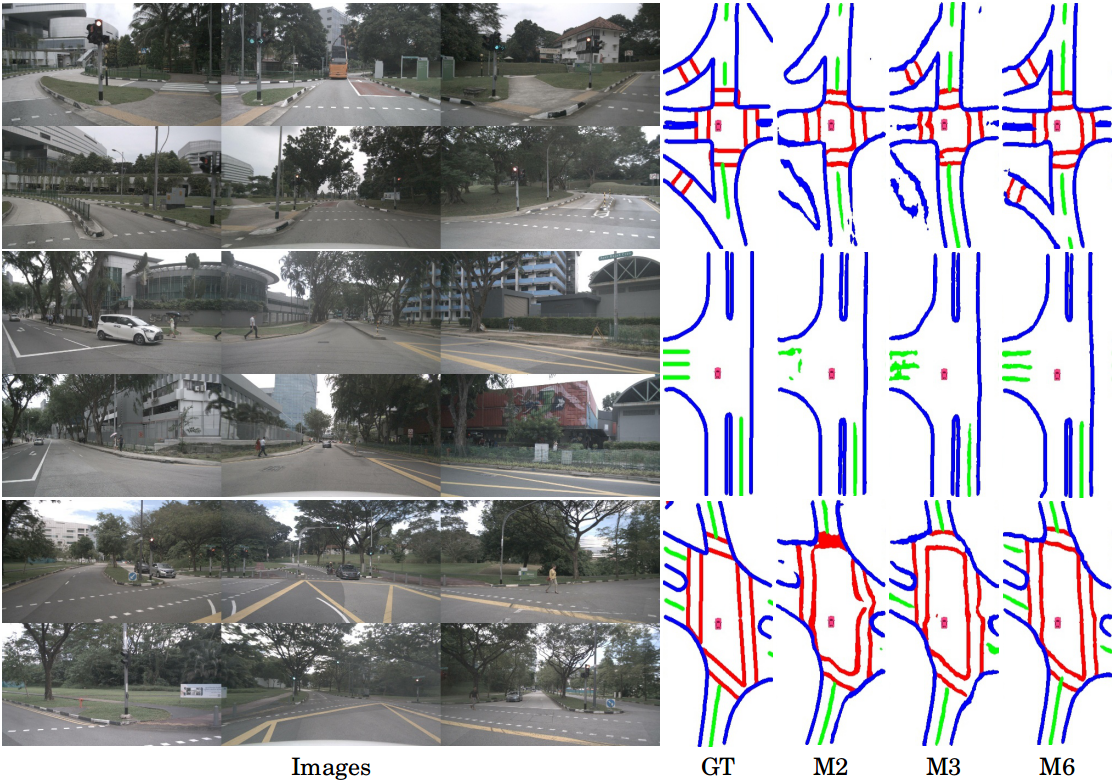}}\\
  \caption{Examples of BEV segmentation results of surrounding cameras on nuScenes val set.} 
  \label{figure:surr_nuscene}
  \vspace{-2mm}
\end{figure}

\subsection{Dataset}

\textbf{nuScenes Dataset.} nuScenes~\cite{caesar2020nuscenes} dataset is a large-scale autonomous driving dataset which consists of 1,000 sequences from 6 surrounding cameras (front left, front, front right, back left, back, back right). In total, it has a training set with 28,130 images and a validation set of 6,019 images. Three classes, lane dividers, lane boundaries and pedestrian crossings are available to evaluate the BEV segmentation. We use all the surrounding cameras and the front camera in the experiments.

\textbf{Nullmax Front Camera Dataset.} 
We collect a dataset from Shanghai highway with a front camera equipped. The dataset includes various scenarios such as crowd traffic, on and off ramp, shadow, lane changing and cut-in. The dataset is divided into 3,905 images for training and 976 images for validation. Traffic lanes are annotated for the evaluation.

\subsection{Experimental Settings}

We conduct experiments on the nuScenes dataset with the same setting of HDMapNet~\cite{li2021hdmapnet}. Ego vehicle localization on HD-map is used to define the region of BEV. By using surrounding cameras, the BEV is set to [-30m,30m] $\times$ [-15m, 15m] around the ego vehicle. Only with the front camera enabled, the BEV area is set to [0m, 60m] $\times$ [-15m, 15m]. Road structure is represented as line segments with 5-pixel width. The groundtruth mask is set to $400 \times 200$. Following STSU~\cite{can2021structured}, we use images of $448 \times 800$ as the input of the network.
Similarly, the BEV area in Nullmax front camera dataset is set to [0m, 80m] $\times$ [-10m, 10m].
The groundtruth mask is of size $512 \times 128$.
Traffic lane is 3-pixel width in the groundtruth mask.
The input image size is $384 \times 640$ on the Nullmax dataset.

We follow deformable DETR~\cite{zhu2020deformable} for the network design. A weighted cross entropy loss of [1, 15, 15, 15] is used in the experiments.
$M = 8$ and $K = 16$ are set for multi-camera deformable attention of  BEV transformer decoder. The embedding dimension in all transformer modules is set to 256, and the feature dimension of FFN module is set to 512. We apply data augmentation including random horizontal flips, random brightness, random contrast, random hue and random swap channels. 
The network is optimized by AdamW~\cite{loshchilov2017decoupled} optimizer with a weight decay of $10^{-4}$. The initial learning rate of the backbone and the transformer is set to $10^{-5}$, $10^{-4}$ and decreased to $10^{-6}$ and $10^{-5}$ at $100$-th epoch. We train models on 4 RTX 3090 GPUs and the per-GPU batch size is 1. All the models are trained from scratch with 120 epochs.

\subsection{Evaluation Metrics and Results}

\begin{figure}[b]
  \centering{\includegraphics[width=0.95\linewidth]{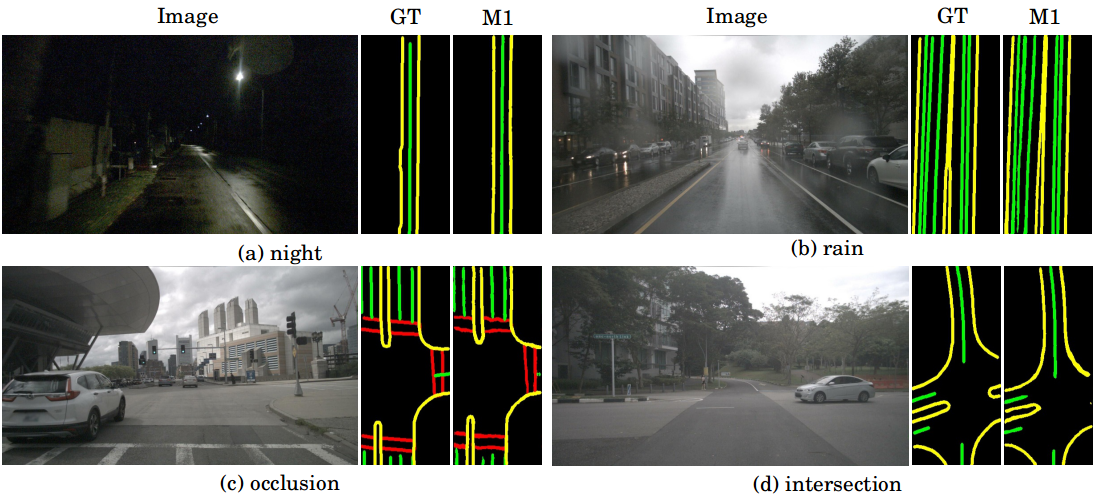}}\\
  \caption{Examples of BEV segmentation results of front camera on nuScenes val set.} 
  \label{figure:fc_nuscene}
\end{figure}

\begin{figure}[t]
  \centering{\includegraphics[width=0.95\linewidth]{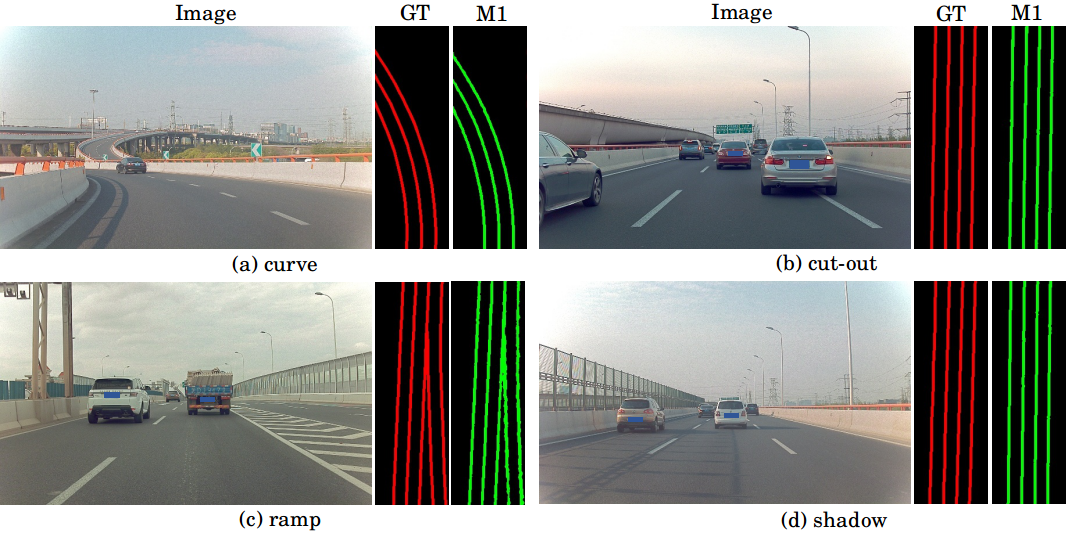}}\\
  \caption{Examples of BEV segmentation results on Nullmax Front Camera val set.} 
  \label{figure:fc_nullmax}
  \vspace{-3mm}
\end{figure}

For quantitative evaluation, we measure the intersection-over-union (IoU) between the segmentation results and the groundtruth. 
IoU and mIoU are  calculated by
\begin{equation}
 { \text{IoU}(S_p, S_g)  = \left | \frac{\left (S_{p}  \bigcap S_{g}\right ) }{\left (  S_{p} \bigcup S_{g}\right )} \right |} ,
\label{iou}
\end{equation}

\begin{equation}
{ \text{mIoU}(S_p, S_g) = \frac{1}{N} \sum_{n=1}^{N} IoU_{n}\left ( S_{p},S_{g}  \right ) } ,
\label{miou}
\end{equation}
where $ S_p \in \mathbb{R}^{H_g \times W_g \times N }$ and $S_g \in \mathbb{R}^{H_g \times W_g \times N }$ are the predicted results and the groundtruth. $H_g$ and $W_g$ denotes the height and width of the groundtruth mask and $N$ is the number of classes in the dataset.

We compare our method with previous state-of-the-art methods on the nuScenes dataset. All these methods do not utilize temporal information for BEV semantic segmentation. The results are summarized in Table.~\ref{tab:sota_nuscene}. 
Experimental results show that our method outperforms the previous state-of-the-art approaches on the nuScenes val set. It surpasses the HDMapNet (surr)~\cite{li2021hdmapnet} by a large margin of $+$10.48 Divider IoU, $+$13.89 Ped Crossing IoU, $+$10.47 Boundary IoU and $+$11.65 All Classes IoU. Segmentation examples are shown in Fig.~\ref{figure:surr_nuscene}.

To investigate that our method has the ability to process arbitrary cameras. We report results only using the front camera on nuScenes dataset in Table.~\ref{tab:fc_nuscene}. Some segmentation results are shown in Fig.~\ref{figure:fc_nuscene}. Only using a front camera, our method also obtains competitive results.  

Similar observation of BEV segmentation on Nullmax front camera dataset is listed in Table.~\ref{tab:fc_nullmax}. Fig.~\ref{figure:fc_nullmax} provides examples of challenging driving scenarios (curve, cut-out, ramp and shadow).

We also explore the impact of the number of queries in the BEV space. The model with 5,000 BEV queries (1/4 of groundtruth mask) achieves better results compared to the model with 1,250 queries. Throughout the experiments, the query number is set to 5,000.

\subsection{Ablations}
In this section, we analyze the effects of the proposed components via an ablation study. 
We study the standard multi-head self attention module, multi-scale deformable self attention module, standard multi-head cross attention module, multi-camera deformable cross attention module and number of layers in the encoder and decoder of our method. Results of various backbones are also investigated. Additionally, the effect of camera position embedding in multi-camera attention is provided. The experimental settings and results are given in Table.~\ref{tab:ablation}. The methods with different protocols are denoted as M1 to M7.

\begin{figure}[t]
  \centering{\includegraphics[width=0.95\linewidth]{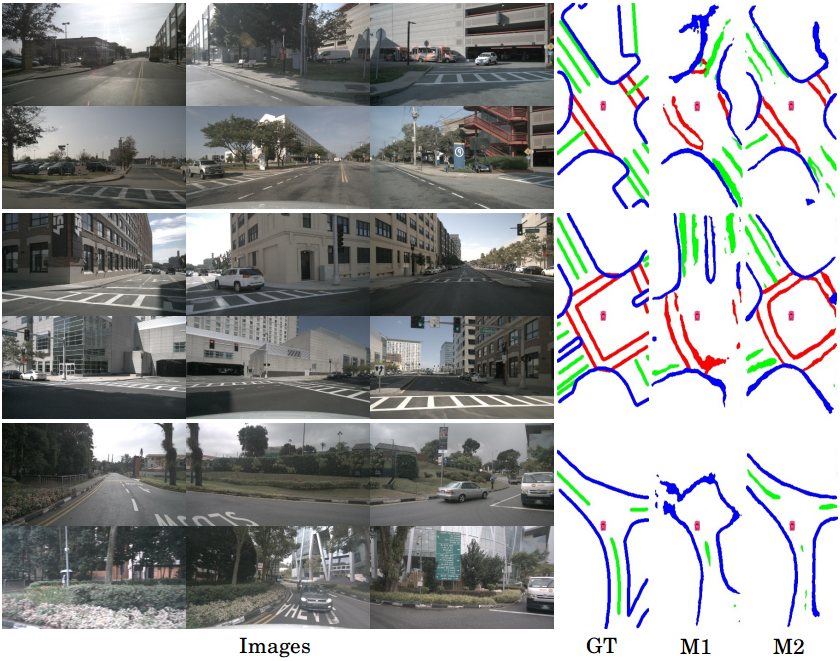}}\\
  \caption{Results of the ablation study analyzing the effects of camera position embedding on nuScenes val set.}
  \label{figure:cam_embed}
  \vspace{-2mm}
\end{figure}

\textbf{Learnable Camera Position Embedding.}

We find that a learnable camera position embedding is important for the standard cross attention module in multi-camera decoder, while it does not provide a positive effect of multi-camera deformable cross attention module.
In Table.~\ref{tab:ablation}, having camera position embedding with standard cross attention module, method M2 improves the segmentation results of M1.
Results in Fig.~\ref{figure:cam_embed} demonstrate that method incorporating camera position embedding obtains superior results at scenarios that the ego vehicle makes a turn at the intersections.
However, by using multi-camera deformable cross attention module, 
M7 with a learnable camera position embedding does not boost semantic segmentation results of M6.

We argue that in standard multi-camera cross module, the position embedding of image features is relative to each camera itself and does not have ability to discriminate different cameras. Therefore, the spatial information of the camera is critical for semantic segmentation tasks using standard multi-camera cross module. In our proposed multi-camera deformable cross attention module, a BEV query learns a separate reference point for each camera. It automatically encode camera position information into the cross attention step. It turns out that an additional and redundant camera position embedding provides marginal effect for the final semantic segmentation accuracy.

\begin{figure}[t]
  \centering{\includegraphics[width=0.9\linewidth]{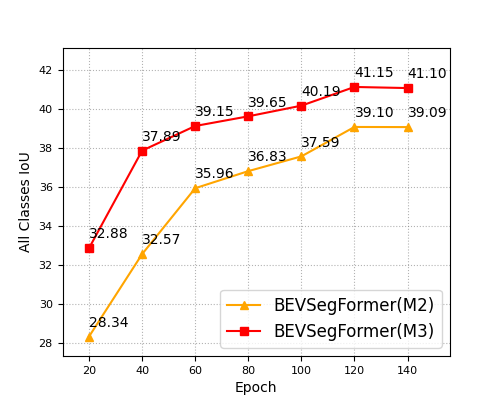}}\\
  \caption{Convergence curves of BEVSegFormer (M2) and BEVSegFormer (M3) on nuScenes val set. The same training schedules is used in both models.} 
  \label{figure:epoch_iou}
  \vspace{-3mm}
\end{figure}

\textbf{Multi-camera Transformer Decoder.} 
Experiments in Table.~\ref{tab:ablation} show that compared to methods M1 and M2 using standard  multi-head cross attention in the decoder, method M3 with multi-camera deformable cross attention in the decoder obtains significant performance gains. Multi-camera deformable decoder does not require to compute the dense attention over feature maps from all the cameras. 
In this way, the learned decoder attention has strong correlation with the BEV segmentation results, which also accelerates the training convergence. Fig.~\ref{figure:epoch_iou} shows the convergence curves of the BEVSegFormer (M3) and BEVSegFormer (M2) models on the nuScenes val set. The former only needs half the training epochs ($epoch=60$) to obtain a higher IoU of all classes than the latter ($epoch=120$).

\begin{figure*}[htb]
  \centering{\includegraphics[width=0.66\linewidth]{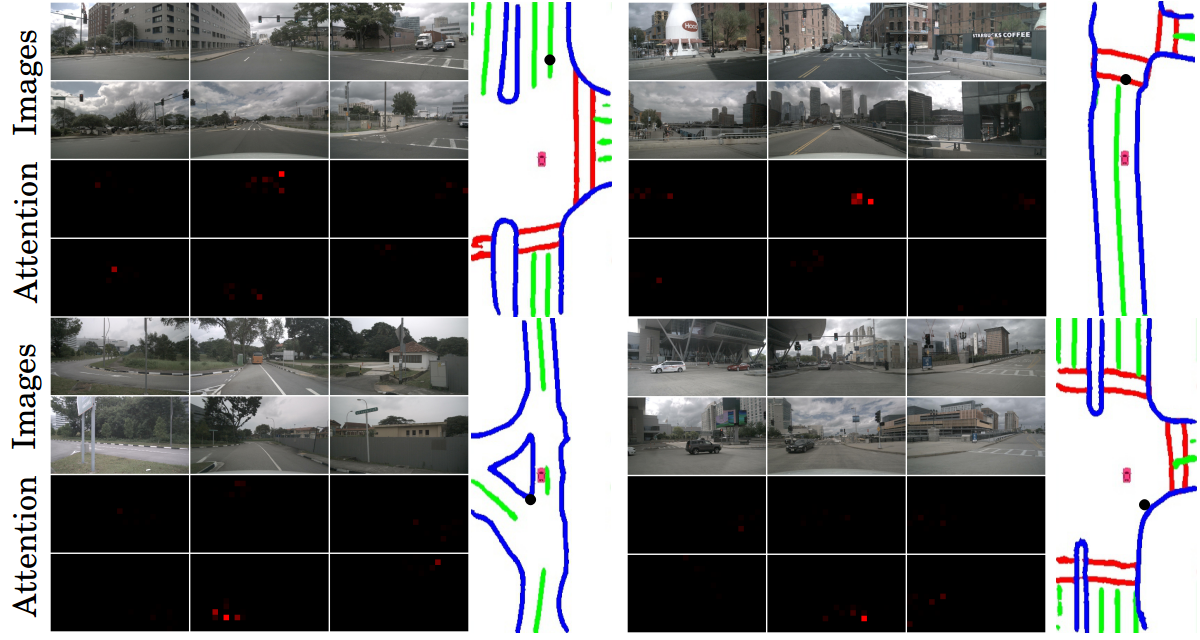}}\\
  \caption{Visualization of Multi-Camera Deformable Cross-Attention weight map in BEV Transformer Decoder module. We visualized the attention weight map of the query corresponding to lane dividers (in green), pedestrian crossings (in red), lane boundaries (in blue) and camera overlap regions on the BEV. Redder pixels indicate larger values of attention weights.} 
  \label{figure:attention_map}
  \vspace{-2mm}
\end{figure*}

In order to investigate the multi-camera deformable cross-attention, we analyze the attention map in Fig.~\ref{figure:attention_map}. Specifically, given a query of road structure on the BEV space, we visualize the corresponding sampling points and attention weights in the last layer of the multi-camera deformable cross-attention. It shows that the sampling points of large weights are clustered near the feature maps of the most relative camera. For example, in the top left of Fig.~\ref{figure:attention_map}, the query (black point) on BEV space is observed in the front camera, its sampling points with large cross-attention weights appear in the feature map of the front camera.  
Even though in some cases that the sampling points are not precisely linked with the desired query, the image features from the relative camera have enough receptive fields to cover the input query on BEV space.

\textbf{Multi-scale Transformer Encoder.} 
In Table.~\ref{tab:ablation},  We also compare the segmentation results using standard transformer encoder (M3) and deformable encoder (M4) which utilizes multi-scale images. Similar to deformable DETR~\cite{zhu2020deformable}, the deformable encoder module not only accelerates the model convergence, but also provides multi-level image features which in turn to boost the BEV segmentation results.

\textbf{Number of Encoder Block and Decoder Block.} 
Table.~\ref{tab:ablation} shows the segmentation results using different numbers of encoder layers and decoder layers. Compared to method M4, the segmentation results of method M5, which has more encoder and decoder layers, are further boosted.

\textbf{Shared Backbone.} Lastly, we study the effect of the backbones. Results of method using ResNet-101 slightly improves the results of method with ResNet-34 as backbone.

\section{CONCLUSIONS}

In this paper, we propose a deformable transformers based method for BEV semantic segmentation from arbitrary camera rigs. Our method adapts the recent detection method, deformable DETR, which processes a single image to BEV semantic segmentation of single or multiple images. We show that our transformer based method outperforms previously reported state-of-the-art approaches. By making use of the deformable BEV decoder, a novel component to parse BEV semantic results from the encoded image features, our method achieves additional boost of the performance. In future work, we would like to explore memory-efficient and more interpretable transformer-based methods for BEV semantic segmentation in autonomous driving.

{\small
\bibliographystyle{ieee_fullname}
\bibliography{refs}
}

\end{document}